# Video Content Classification using Deep Learning

Saraswati Patil[1], Pradyumn Patil(11911229), Vishwajeet Pawar(11911075), Yashraj Pawar(11911350), Shruti Pisal(11910910)

Department of Computer Engineering

***Abstract*** — *Video content classification is an important research content in computer vision, which is widely used in many fields, such as image and video retrieval, computer vision. This paper presents a model that is a combination of Convolutional Neural Network (CNN) and Recurrent Neural Network (RNN) which develops, trains, and optimizes a deep learning network that can identify the type of video content and classify them into categories such as "Animation, Gaming, natural content, flat content, etc". To enhance the performance of the model novel keyframe extraction method is included to classify only the keyframes, thereby reducing the overall processing time without sacrificing any significant performance.*

I. INTRODUCTION

Video has become more popular in many applications in recent years due to increased storage capacity, more advanced network architectures, as well as easy access to digital cameras, especially in mobile phones. Today people have access to a tremendous amount of video, both on television and the Internet. The amount of video that a viewer has to choose from is now so large that it is infeasible for a human to go through it all to find a video of interest. One method that viewers use to narrow their choices is to look for videos within specific categories or genres. Because of the huge amount of video to categorize, research has begun on automatically classifying video. , classification and analysis of videos. For this reason, it is necessary to have a system for generating relevant labels to a video or different parts of the video automatically without human intervention. The video content should be extracted and understood from the video data to find the type of video it belongs to. In this project, we will create a classifier for video content.It will extract the keyframes from a given video and will feed these frames to the CNN plus RNN classifier which will label it to a particular class i.e animation, news ,games, scenery, etc

II. LITERATURE REVIEW

Over the last decade, active researchers have produced methods for improving Video Classification..
In [1], this paper gives a method to reduce the computation time for video classification using the idea of distillation. Specifically, we first train a teacher network that computes a representation of the video using all the frames in the video. The proposed approach in [2] used an improved CNN-based classification algorithm with a softmax loss function to classify mine video scenes. Compared to the older algorithms this CNN algorithm showed higher accuracy, precision, and recall in classifying mine videos. It did solve the drawbacks that were there in the traditional model but as it requires massive data to give a more accurate result which was not available at that moment so it was predicted that future massive data and complex training would make that model more efficient.The presented method in paper [3] considers the problem
constructing a minimum representative set of video
physical features that can be used in different regression
and classification tasks when working with video. To test the
the resulting set of features, the problem of classifying video
into 4 classes was considered: cartoon, drone shooting,
computer game and sports broadcast.
[4] uses support vector machine.Proposed
The system is capable of extracting keyframes from videos and classifying them as natural scenery, personality, animal and plant.For feature extraction, new methods like edge histogram , dominant color, color layout and face feature were used for image retrieval.Results have shown a high accuracy when parameters are C=14 and y =0.4.this research [5 ]paper proposes a novel method for sports video scene classification with the particular intention of video summarization.In this paper discussed methodologies in the recent past for shot classification exclusively or part of their scheme, conclusively, deep-learning approaches presented high quality of the classification. Deep learning proved its importance in image classification, and incorporating the pre-trained model and applying augmentation .[6] In this paper, a template-based keyframe extraction method is proposed which employs action template-based similarity to extract keyframes for video classification tasks.Combining pre-trained CNN with ConvLSTM has achieved the highest classification accuracy among the other architectures.One of the limitations is that CNN architecture used did not produce the best results .Also it requires a machine with more than one GPU .So future work could be focused the on application of the proposed algorithm using more powerful architectures for real-world video classification.

[1]

III. METHODOLOGY/EXPERIMENTAL

*1. Dataset*

The COIN dataset consists of 11,827 videos related to 180 different tasks, which were all collected from YouTube. The average length of a video is 2.36 minutes. Each video is labelled with 3.91 step segments, where each segment lasts 14.91 seconds on average. In total, the dataset contains videos of 476 hours, with 46,354 annotated segments.

We have considered 5 classes out of the given 12 classes and have taken 1000 images for training and testing purpose

*2. Keyframe Extraction*

Key-frames are defined as the representative frames of a video stream, the frames that provide the most accurate and compact summary of the video content.

Frame extraction and selection criteria for key-frame extraction

- Frame that are sufficiently different from previous ones using absolute differences in LUV color space
- Brightness score filtering of extracted frames
- Entropy/contrast score filtering of extracted frames
- K-Means Clustering of frames using image histogram
- Selection of best frame from clusters based on and variance of laplacian (image blur detection)

*3. Video Classification Model*

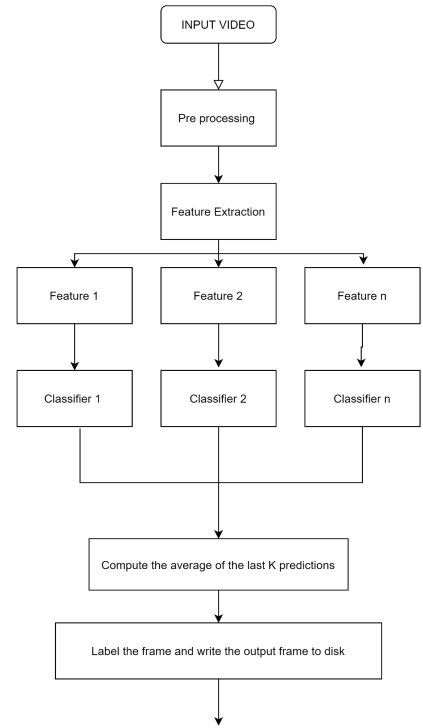

Suppose we have $n$ number of classes, let's name them $\{c_1, c_2, c_3, \ldots c_n\} \in C$ and '$S$' video containing '$M$' number of frames present in it as $\{F_0, F_1, F_2, \ldots F_{M-1}\}$. The purpose of this model is to classify the frames by identifying the classes they belong to.

For each of the unique '$n$' number of classes we have we will be predicting the probability $P(c_n|S)$

$$P(c_n|S) = \kappa(\{F_0, F_1, F_2, \ldots F_{M-1}\}).$$

The mentioned $\kappa$ can be a Neural Network that does the job of predicting the frames '$F$'.

Now we discuss in detail about the model architecture we are going to implement this neural network.

3.1. CNN architecture(spacial)

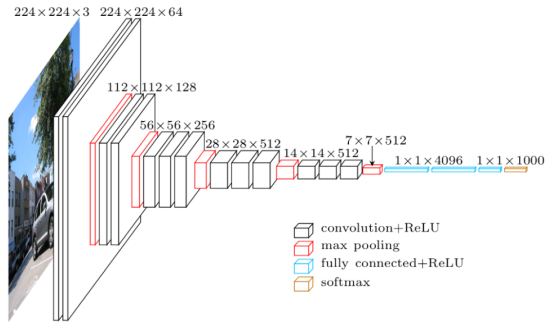

*Fig. 1*

Neural networks are considered a very powerful and intelligent technique to classify different types of data for important applications. New types of neural networks called convolutional neural networks (ConvNets or CNNs).were developed classification and recognition tasks for many applications. It makes a breakthrough in image processing, video, speech, and text recognition research, and the concentration it earned is due to the capability and the high performance.

RNN (temporal)

➔ CNN +RNN - Classification and training.

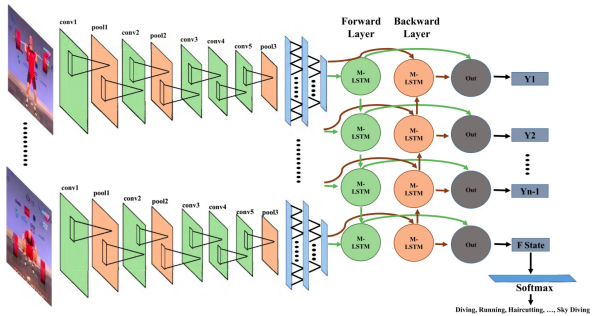

As we are working on videos we need to look at both spatial as well as temporal factor to make sure system understands the coso we need a hybrid model for classification of video.And we know that CNN i.e. Convolutional neural networks can take in an input image, assign importance to various aspects/objects in the image and be able to differentiate one from the other which means it can work on the spatial features of the video while RNN i.e Recurrent Neural networks which uses sequential data or time-series data so it can work on the temporal features.

➔ Feature extraction :

➔ Convolution Layer
It contains a set of digital filters to perform the function of convolution in the input data.It is the number of filters the convolutional layers will learn from. It is the absolute value and determines the number of filters that come out of convolution. It uses the 'Relu'activation function here.

➔ Average Pooling Layer
It involves calculating the average for each part of the feature map. This means that each 2×2 square of the feature map is down sampled to the average value in the square

➔ Max Pooling Layer
It is an operation that calculates the maximum value for patches of a feature map, and uses it to create a downsampled (pooled) feature map.It is used to reduce the dimensions of the feature maps

➔ DropOut Layer
It is a technique used to prevent a model from overfitting.It works by randomly setting the outgoing edges of hidden units (neurons that make up hidden layers) to 0 at each update of the training phase.

➔ Dense Layer
We can call it a deeply connected network. Each neuron in the dense layer receives input from all neurons of its previous layer.Acts like output layer and givas results according to classified classes.

➔ SoftMax Layer
It assigns decimal probabilities to each class in a multi-class problem. Those decimal probabilities must add up to 1.0. This additional constraint helps training converge more quickly than it otherwise would.It is implemented through a neural network layer just before the output layer

3.2. Workflow

➔ Firstly the dataset is loaded
➔ The images frames are resized to a common scale of 224 x224.
➔ As we do not have any separate testing dataset we will be dividing the current dataset into training and validation dataset. in the ratio of 80:20

- ➔ The datasets are loaded with a batch size of 128 and then the dataset is shuffled to avoid biased conditions.
- ➔ And the class mode will be categorical.
- ➔ Then the images are preprocessed for training using feature extraction and edge detection methods.
- ➔ The above process gives a compiled model so from here the model has to be fit to train. So the model is trained over the training dataset under 15 epochs.
- ➔ For the Testing Part , Video input is taken and the keyframes are then fed in the pre-trained model.where the input frame image of size 224 x 224 is given to the model.
- ➔ The result of prediction is given in form of percentage where it tells that by what percentage does the video resembles each class

### IV. RESULTS AND DISCUSSIONS

The model after compilation is trained and then tested after this process the model has given an accuracy of 80.27%. This accuracy was possible because of introducing some complex layers like the Dense layer and Dropout layer where the random neurons are dropped for training. Also as we used the softmax function which helped in quicker training coverage. Current results of can be compared with the InceptionV3 model developed by google .

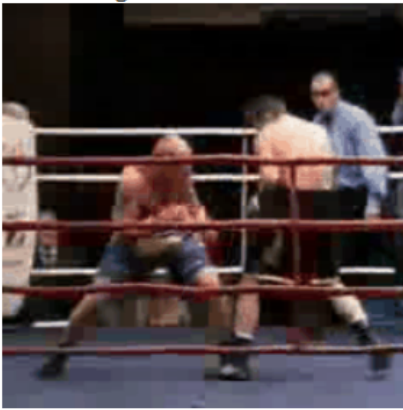

Fig . 2

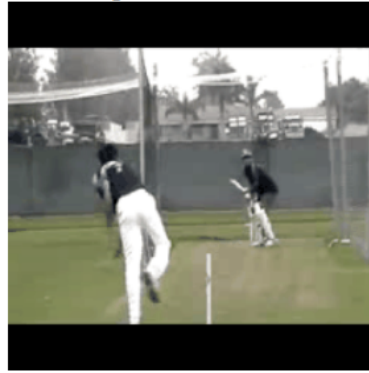

### V. LIMITATIONS

- ➔ The Processing was carried out on low-end hardware resulting in less processing power therefore a small subsample of Dataset was used to conduct the tasks.
- ➔ The absence of a dedicated 'GPU' makes the computational task a bit time-consuming.

### VI. FUTURE SCOPE

- ➔ Use the model for more specific purposes related to '*video classification*' such as '*Action Recognition*' , '*Video Caption generator* ' 'Event Detection in a video 'and etc.
- ➔ Use container-orchestration system Tools such as 'Docker' and 'Kubernetes' for easy deployability and scalability.
- ➔ Integrate Cloud processing platforms such as 'Google cloud' to increase the efficiency of the overall system.

### VII. CONCLUSION

By comparing the obtained results with the result generated from the Inception V3 model . the presented model is able to emulate the accuracy of approximately 80.27 % as expected.Further Processing using the specified "Coin Dataset" is to be done for further evaluation. This accuracy was possible because of introducing some complex layers like the Dense layer and Dropout layer where the random neurons are dropped for training. Also as we used the softmax function which helped in quicker training coverage

## VIII. Acknowledgment

We would Like to thank and express our gratitude to "Prof. Saraswati Patil " for Her guidance and support in our project. This research was supported/partially supported by the Vishwakarma Institute of Technology Pune. We thank our colleagues for providing insight and expertise that greatly assisted the research, although they may not agree with all of the interpretations/conclusions of this paper.We thank [Prof. Saraswati Patil for assistance with image processing and deep learning concepts. S.T. Patil for comments that greatly improved the manuscript. We are also immensely grateful to our H.O.D Sagar Shinde for their comments on an earlier version of the manuscript, although any errors are our own and should not tarnish the reputations of these esteemed persons.